\pdfoutput=1
\documentclass[11pt]{article}

\usepackage[preprint]{acl}

\usepackage[utf8]{inputenc}
\usepackage[T1]{fontenc} %% for \textbackslash
\usepackage[cjk]{kotex}
\usepackage{CJKutf8}

\usepackage{times}
\usepackage{latexsym}

\usepackage[T1]{fontenc}
\usepackage[utf8]{inputenc}

\usepackage{microtype}

\usepackage{synttree}
\usepackage{subcaption}
\usepackage{caption}

\usepackage{inconsolata}
\usepackage{amsmath} 
\usepackage{tabularx}
\usepackage{arydshln}
\usepackage{graphicx}

\title{Revisiting Absence with \textit{Symptoms that *T* Show up Decades Later} to Recover Empty Categories}
\author{
  Emily Chen$^{1*}$ $\quad$ Nicholas Huang$^{1*}$ $\quad$ Casey Robinson$^{2*}$ $\quad$ Kevin Xu$^{1}$\thanks{Equally contributed authors} \\
  \large{\textbf{Zihao Huang}}$^{2}$ $\quad$ \large{\textbf{Jungyeul Park}}$^{2}$ \\
  $^{1}$ Department of Computer Science $\quad$ $^{2}$ Department of Linguistics \\
  The University of British Columbia\\
  Vancouver, BC, Canada \\
  \texttt{\{emilyec0,nhuang05,crobin04,kxu20,zihowong\}@student.ubc.ca} $\quad$ \texttt{jungyeul@mail.ubc.ca}\\ 
}

\begin{document}
\maketitle 
\begin{abstract}
This paper explores null elements in English, Chinese, and Korean Penn treebanks. Null elements contain important syntactic and semantic information, yet they have typically been treated as entities to be removed during language processing tasks, particularly in constituency parsing. 
Thus, we work towards the removal and, in particular, the restoration of null elements in parse trees. We focus on expanding a rule-based approach utilizing linguistic context information to Chinese, as rule based approaches have historically only been applied to English. We also worked to conduct neural experiments with a language agnostic sequence-to-sequence model to recover null elements for English (PTB), Chinese (CTB) and Korean (KTB). To the best of the authors’ knowledge, null elements in three different languages have been explored and compared for the first time. In expanding a rule based approach to Chinese, we achieved an overall F1 score of 80.00, which is comparable to past results in the CTB. 
In our neural experiments we achieved F1 scores up to 90.94, 85.38 and 88.79 for English, Chinese, and Korean respectively with functional labels. 
% In our neural experiments we achieved F1 scores of 90.94/88.67 for English with/without labels, 83.76/85.38 for Chinese with/without labels, and 88.98/88.79 for Korean with/without labels. 

\end{abstract}

\section{Introduction}

Null elements, or empty categories, are elements that exist within a parse tree that do not correspond to a surface level word. 
Chomsky outlined how certain elements could be \textit{invisible} or \textit{null} in surface structure \citep{chomsky:1981}. These null elements serve an important role in syntax, especially in explaining movement and transformational grammars \citep{chomsky:1965}.
In the context of dependency parsing, particularly in the context of predicate elision, null elements have predominantly surfaced in the enhanced representation of universal dependencies.\footnote{\url{https://universaldependencies.org/u/overview/enhanced-syntax.html}} Such elements can encode additional semantic information of sentences by recording non-local dependencies, enabling interpretation of constructions such as WH-questions \citep{johnson-2002-simple}. 
Conversely, within constituency parsing, null elements have typically been treated as preprocessed entities to be removed \citep{collins:1999,bikel:2004:CL}.

However, excluding null elements can be problematic when attempting to parse more than just the syntactic information from a surface sentence, since the semantic information they provide is often lost. Certain languages, such as Japanese, Korean, and Chinese, more readily omit specific arguments within sentences compared to English when these arguments have been previously referenced or can be inferred. As an example, when attempting machine translation from pro-drop languages such as Chinese and Korean, where pronouns are often dropped, into English, problems arise because English pronouns must be inferred from non-existent words to preserve meaning \citep{chung-gildea-2010-effects}. Therefore, recovering null elements from a parsed tree without traces may enable us to resolve null anaphora, where omitted pronouns or other referential expressions are understood from context. Outside of practical applications, null elements have a rich history in syntax, and are an essential aspect of explaining transformational grammars. By restoring null elements, parse trees may better reflect a linguistic understanding of movement, as well as many other linguistic phenomena relevant to the use of null elements.

Previous work has approached the recovery of null elements in different ways, ranging from more rule based approaches to statistical approaches. For example,
\citet{johnson-2002-simple} introduced a pattern-matching algorithm aimed at recovering empty nodes within the Penn English Treebank (PTB) \citep{marcus-santorini-marcinkiewicz:1993:CL}. Their exploration involved scrutinizing parsing outcomes for various null elements such as *T* (WH trace) and *U* (empty units), demonstrating their potential to enhance parsing results. \citet{campbell-2004-using} proposed a linguistically oriented approach grounded in Government-Binding theory \citep{chomsky:1981}, which forms the basis of syntactic annotation in treebanking. Despite the inclination towards a rule-based method for empty category retrieval, reported parsing outcomes have been less compelling \citep[p.650]{campbell-2004-using}.
Pattern matching algorithms have also been utilized to restore Restoring null elements for Chinese and Korean using a similar algorithm to Johnson, but included more context in the generated patterns to better recover null elements in Chinese, since Chinese contains null elements (*PRO* and *pro*) which have similar syntactic contexts so need wider patterns to be effectively recovered. \citep{chung-gildea-2010-effects}. 
In contrast, \citet{levy-manning:2004:ACL}  introduced an algorithm based on statistical features for reconstructing non-local dependencies concerning null elements. Non-local dependencies are syntactic constructs where two words cannot be obviously related based on typical syntactic rules -- null elements that are indexed to the dependant word are a way to represent non-local  dependencies using existing syntactic rules.  

Previous work on null element restoration has primarily focused on English -- we hope to expand the scope of studies by implementing algorithms for null element recovery for English, Chinese, and Korean. Furthermore, we aim to take a a comparative approach by implementing linguistically-motivated algorithms for removing and restoring null elements, as well as expanding to a language agnostic neural experiments. Considering the general success of neural experiments in outperforming traditional statistical parsers, applying neural experiments to null element recovery may allow for greater accuracy than previous approaches.

\section{Previous Work}
After being introduced by Chomsky, null elements have become a standard feature of syntax and have been a topic of further exploration. \citet{reeves-1992-zero} discusses in depth the legitimacy and theory behind null elements in a purely theoretical context by generating a framework in which null elements are not allowed and testing if such a framework fails.
Reeves concludes that while sometimes disruptive to work with, null elements are ultimately necessary for systematically representing and analyzing the English language.

Previous works on the computational approach to restoring null elements generally choose one of three methods for inserting null elements; pre-processing, in-processing, and post-processing. Such previous works and their respective approaches are noted in Table~\ref{previous-work-comparison}. We utilize the post-processing approach. In this section we give a broad overview of each approach and justify our decision to use post-processing.

As noted in Table~\ref{previous-work-comparison}, \citet{dienes-dubey-2003-antecedent} inserted null elements as a step before parsing (thus, preprocessing). However, preprocessing approaches to null element insertion have been linked to worse results for general parsing than with post-processing approaches \citep{johnson-2002-simple}. Advantages of pre-processing versus post-processing or in-processing include runtime efficiency. \citep{dienes-dubey-2003-antecedent}.

Other work utilized in-processing, making this the most popular approach from previous attempts of inserting null elements \citep{dienes-dubey-2003-deep,schmid-2006-trace,kato-matsubara-2016-transition,hayashi-nagata-2016-empty,kummerfeld-klein-2017-parsing}. In in-processing, null elements are integrated into the parsing algorithm. Generally, in-processing results from previous work are relatively similar to post-processing results. However, in-processing is a more complex task than post-processing, and is less conducive to a rules based approach, since a rule-based approach to null element recovery uses existing syntactic structure to predict the position of null elements. 

Finally, \citet{johnson-2002-simple} and \citet{campbell-2004-using} utilized post-processing, during which null elements are inserted into the parse tree. Results from post-processing are favourable to some in-processing results \citet{campbell-2004-using}. Furthermore, post-processing avoids problems where the null element is inserted in the correct position, but there maybe bracketing errors higher up in the tree, such as in the grandparent of the null element. Preventing these types of errors simplifies the evaluation process, since we prevent the need for assessing the overall syntactic structure and can evaluate just the position and type of the null elements.

For this paper, we implement a post-processing approach \citep{johnson-2002-simple,campbell-2004-using}. We believe that this approach might be the best for the insertion of null elements, as it has yielded promising results in previous studies without adding further complexities. However, this method has not been tested extensively in recent years, especially with neural approaches, including a language-agnostic sequence-to-sequence model.

The Berkeley Neural Parser has historically been used in processing trees from the Penn Treebank, and can be used to produce trees with and without traces. In trees without traces, null elements are thus removed. This approach works very well for English and Chinese to achieve the goal of removing null elements, but there is a knowledge deficit as there is no parser to remove null elements in Korean sentences. Furthermore, there have been no previous works attempting to restore null elements into Korean sentences. To address this knowledge gap, we have chosen to build off of the approach used by the Berkeley Neural Parser to similarly parse and remove null elements from Korean sentences, as well as use neural experiments to attempt the restoration of null elements into Korean sentences.

\begin{table}[!ht]
    \centering
% \resizebox{.48\textwidth}{!}
{\footnotesize
\centering
\begin{tabular}{r | c  } \hline 
 % &  &  \\ \hline  
\citep{johnson-2002-simple} & post-processing     \\
\citep{dienes-dubey-2003-deep} & in-processing \\
\citep{dienes-dubey-2003-antecedent} & pre-processing \\
\citep{campbell-2004-using} & post-processing    \\

\citep{schmid-2006-trace} & in-processing \\
\citep{kato-matsubara-2016-transition} & in-processing \\
\citep{hayashi-nagata-2016-empty} & in-processing \\
\citep{kummerfeld-klein-2017-parsing} & in-processing \\\hline 
\end{tabular}
}
    \caption{Previous work approach comparison}
    \label{previous-work-comparison}
\end{table}

\section{Null Elements Typology}
% \jp{REMOVE all types only for Korean, and revise tables..}
% \jp{THIS SECTION IS TOO LONG... (for now keep it as it is); but if we want a LONG PAPER (8pages) the current status is okay (If we go with a long paper, it should a DEEP analysis of experiment results). TRY TO DESCRIBE what are common and different between English and Chinese.}
The Penn Treebank contains sentences sourced in English \citep{marcus-santorini-marcinkiewicz:1993:CL}, Chinese \citep{xue-EtAl:2005} and Korean \citep{han-EtAl:2002} -- all of which contain null elements that work to communicate implicit syntactic structures. Table~\ref{null-elements-in-languages} in the appendix below shows which types of null elements appear in each language (English, Chinese and Korean). From this table we can see that English does not share many null elements with pro-drop languages such as Chinese and Korean; only traces of A' movement (*T*) are present in all languages. Below, we also provide a brief description of each null element type -- for further detail please refer to the Penn tree bracketing guidelines.

% \begin{table}[!ht]
% \centering
% \resizebox{.49\textwidth}{!}
% {\footnotesize
% \begin{tabular}{c|ccc} \hline 
%      &  English & Chinese & Korean \\ \hline 
% Trace of A' Movement (*T*)     & O & O & O \\ 
% Trace of NP movement (NP *)    & O & X & X \\
% Empty Units (U)                & O & X & X \\
% Null Complementizer (0)        & O & X & X \\
% Control Constructions (*PRO*)  & X & O & X \\
% Pro-drop (*pro*)               & X & O & O \\
% Null operator (*op*)           & X & O & O \\
% Predicate Deletion (*?*)       & X & X & O \\
% Right Node Raising (*RNR*)     & O & O & X \\
% \hline 
% \end{tabular}
% }
% \caption{Whether a null element appears in English, Chinese, or Korean. Cells are marked with O if that null element does appear in the language, and X if not.} \label{null-elements-in-languages}
% \end{table}

\begin{table}[!ht]
\centering
\resizebox{.48\textwidth}{!}
{\footnotesize
\begin{tabular}{c|ccc} \hline 
     &  English & Chinese & Korean\\\hline  
Trace of A' Movement (*T*)     & O & O  & O \\ 
Trace of NP movement (NP *)    & O & X  & X \\
Empty Units (U)                & O & X  & X \\
Null Complementizer (0)        & O & X  & X \\
Control Constructions (*PRO*)  & X & O  & X \\
Pro-drop (*pro*)               & X & O  & O \\
Null operator (*op*)           & X & O  & O \\
Predicate Deletion (*?*)       & X & X  & O \\
Right Node Raising (*RNR*)     & O & O  & X \\
\hline 
\end{tabular}
}
\caption{Whether a null element appears in English, Chinese or Korean. Cells are marked with O if that null element does appear in the language, and X if not.} \label{null-elements-in-languages}
\end{table}

\paragraph{Trace of A' Movement (*T*)}
T* is broadly produced via movement, and can be thought of as marking the interpretation location of other constituents that have been moved out of their usual position. Because *T* elements correspond to the interpretive location of some other constituent in the sentence, *T* always carries a referential index to the constituent in the sentence to which it corresponds. *T* is common in relative clauses across the three languages, but Chinese also commonly uses the *T* element in the context of topicalization. 
In Korean, *T* usually results from fronting an argument in a position before its subject. 

\paragraph{Trace of NP movement, Controlled Pro, Arbitrary Pro (*)}
In English, the null element (NP *) is used in constructions of the passive, in which case it appears as the object of a verb or preposition, or in other cases where an NP is elided such as in imperative clauses or infinitive constructions. (NP *) can occur either with a definite reference, or can be arbitrary, in which case it will not be co-indexed. In contrast, Chinese and Korean do not have a clear passive form. The closest counterpart in Chinese is bei-constructions where (NP *) is not usually needed to analyze. Although some sentences in Chinese can be parsed to include (NP *) (such as in short bei-constructions) the data-sets we used contained no such sentences, so, for the sake of this study we will assume that (NP *) appears in English only. 
%An example sentence is shown in Figure 3. 

\paragraph{Empty Units (U) }
Empty units are marked with U and are used in English sentences, and are included when the units of some number are elided. 
%An example sentence utilizing empty units is given in Figure 4. 

\paragraph{Null Complementizer (0)}
Null complementizers are marked with `0' and are used in English when a complementizer is elided, such as in the sentence, "I know (that) he is coming," where the complementizer can be elided, leaving a null element.

\paragraph{Control Constructions (*PRO*)}
*PRO* does not occur in English sentences in the Penn Treebank, but does occur in Pro-drop languages such as Korean and Chinese. *PRO* is a null element that can only occur as the subject of a sentence, and occurs in two general forms. In one such case, the *PRO* will have a variable reading and will appear as the main subject of the matrix clause. It can also appear with a definite reference, in which case it is usually the subject of an embedded clause. *PRO* is distinguished from *pro* in that *PRO* is in complementary distribution with overt subjects, so sentences containing *PRO* cannot replace *PRO* with an overt NP and remain grammatical.
%An example containing *PRO* is provided in Figure 7.  

\paragraph{Pro-drop situations (*pro*)}
The Null element *pro* also occurs in the Pro-drop languages Chinese and Korean. As described above *pro* is similar to *PRO,* but contrasted in the fact that they can be replaced by an overt NP. 
%An example for pro-drop situations is given in Figure 5.

\paragraph{Null Operators (*OP*)}
Null operators are used in Chinese and Korean sentences to form relative constructions. Constructions involving *OP* also leave traces later in the sentence, marked by *T,* thus, *OP* is always co-indexed with a later *T.* *OP* is interpreted as a null WH word, either WHPP or WHNP depending on whether the null trace appears as a adjunct or subject/object. 
%Refer to the Chinese sentence in Figure 2 for an example sentence containing a null operator. 

\paragraph{Right node raising (*RNR*)}
In English, right node raising has now been categorized as an illegal null element by the Penntree Guide (97). However, (*RNR*) was found to be present in provided datasets (such as in LDC99T42), likely since this dataset was constructed before changes that removed (*RNR*) were made. In the context of English, we will therefore consider (*RNR*) in the context of null element removal, but will not work to restore null elements marked as (*RNR*). 
*RNR*, also appears readily in Chinese, and continues to be used in current interpretations. We therfore will seek to restore *RNR* in Chinese. 

\paragraph{{Predicate Deletion (*?*)}}
The symbol *?* is used in coordinated structures in Korean, since the predicate in the second argument of a coordinated structure can be deleted on the assumption that they refer to an earlier predicate. *?* has a similar role and distribution to *RNR* as used in English and Chinese. 

% \subsection{Frequency of Different Null Element Types}
In Table~\ref{null-elements-count}, the frequency of each null element described previously in each language is documented.

\begin{table*}[!ht]
    \centering
% \resizebox{\textwidth}{!}
{\footnotesize
    \begin{tabular}{c|ccccccccc | c  } \hline 
     & *T* & * & *U* & 0 & *PRO* & *pro* & *OP* & *RNR* & *?* & Total \\ \hline
English  & 618 & 515 & 364 & 373 & - & - & - & 12 & - & over 1700 sentences \\
        & 0.36 & 0.30 & 0.21 & 0.22 & - & - & - & 0.0071 & - & per sentence \\ \hline
Chinese & 133 & - & - & - & 57 & 47 & 132 & 10 & - & over 352 sentences\\
   & 0.38 & - & - & - & 0.16 & 0.13 & 0.29 & 0.028 & - & per sentence\\ \hline
Korean  & 736 & - & - & - & - & 950 & 656 & - & 11 & over 535 sentences \\
   & 1.38 & - & - & - & - & 1.78 & 1.23 & - & 0.021 & per sentence \\\hline 
    \end{tabular}
}

\caption{Top: Count of each type of null element in each language. Bottom: Ratio of null element to total number of sentences}  \label{null-elements-count}
\end{table*}

\section{Removing Null Elements}
\begin{figure}
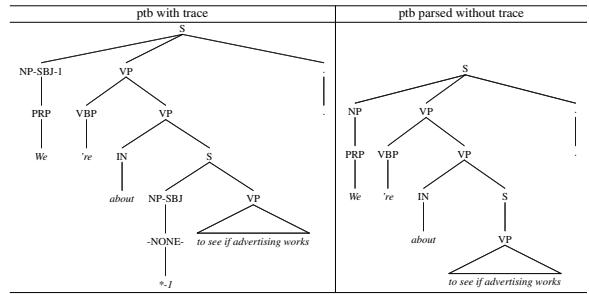

\centering
\resizebox{.48\textwidth}{!}{
\begin{tabular}{c | c} \hline 
{ptb with trace}     &  {ptb parsed without trace} \\ \hline 
\footnotesize{ 
\synttree
[S [NP-SBJ-1 [PRP [\textit{We}]]] [VP [VBP [\textit{'re}]] [VP [IN [\textit{about}]] [S [NP-SBJ [-NONE- [\textit{*-1}]]] [VP [.x \textit{to see if advertising works}]]]]] [$\cdot$ [$\cdot$]]]
}
& 
\footnotesize{
\synttree
[S [NP [PRP [\textit{We}]]] [VP [VBP [\textit{'re}]] [VP [IN [\textit{about}]] [S [VP [.x \textit{to see if advertising works}]]]]] [$\cdot$ [$\cdot$]]]
} \\ \hline 
\end{tabular}
}
\caption{English (PTB), with and without traces}
\label{ptb-examples}
\end{figure}

\begin{figure}
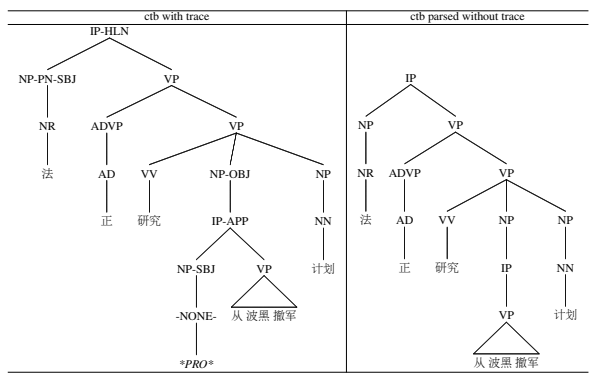

\centering
\resizebox{.48\textwidth}{!}{\footnotesize
\begin{tabular}{c | c} \hline 
{ctb with trace}     &  {ctb parsed without trace} \\ \hline 
\footnotesize{ 
\synttree
[IP-HLN [NP-PN-SBJ [NR [\begin{CJK}{UTF8}{gbsn}法\end{CJK}]]] 
[VP [ADVP [AD [\begin{CJK}{UTF8}{gbsn}{正}\end{CJK}]]] 
[VP [VV [\begin{CJK}{UTF8}{gbsn}{研究}\end{CJK}]] [NP-OBJ [IP-APP [NP-SBJ [-NONE- [\textit{*PRO*}]]] 
[VP [.x \begin{CJK}{UTF8}{gbsn}{从 波黑 撤军}\end{CJK}]]]]
[NP [NN [\begin{CJK}{UTF8}{gbsn}计划\end{CJK}]]]]]]
}
% [VP [PP-DIR [P [\begin{CJK}{UTF8}{gbsn}从\end{CJK}]] [NP-PN [NR [\begin{CJK}{UTF8}{gbsn}{波黑}\end{CJK}]]]] [VP [VV [\begin{CJK}{UTF8}{gbsn}{撤军}\end{CJK}]]]]]
& 
\footnotesize{
\synttree
[IP [NP [NR [\begin{CJK}{UTF8}{gbsn}法\end{CJK}]]] [VP [ADVP [AD [\begin{CJK}{UTF8}{gbsn}{正}\end{CJK}]]] [VP [VV [\begin{CJK}{UTF8}{gbsn}{研究}\end{CJK}]] [NP [IP 
% [VP [PP [P [\begin{CJK}{UTF8}{gbsn}从\end{CJK}]] [NP [NR [\begin{CJK}{UTF8}{gbsn}{波黑}\end{CJK}]]]] [VP [VV [\begin{CJK}{UTF8}{gbsn}{撤军}\end{CJK}]]]]] 
[VP [.x \begin{CJK}{UTF8}{gbsn}{从 波黑 撤军}\end{CJK}]]]]
[NP [NN [\begin{CJK}{UTF8}{gbsn}计划\end{CJK}]]]]]]
} \\ \hline 
\end{tabular}
}
\caption{Chinese (CTB), with and without traces}
\label{ctb-examples}
\end{figure}

% \begin{enumerate}
%     \item \jp{IMPORTANT: ESTABLISH a pipeline (a script) for removing null elements for English and Chinese. DESCRIBE what are common and different between English and Chinese when we remove null elements. ** A README file in Appendix how to remove null elements. }
% \item \jp{we do not write a "technical" report (Therefore, do not ment what the berkeley neural parser is doing).  for removing null elements, we should describe how we "remove" null elements, say if a non-terminal has only a null element as a child, this non-terminal should also be removed... most importantly, we should mention how tree structures are changed after removing null elements. }
% \end{enumerate}
The removal of null elements has, in the past, been used as a step in prepossessing, since null elements have often primarily been seen as entities to remove. In particular, the Berkeley Neural Parser has been a useful tool in processing trees from the Penn Treebank. The Berkeley Neural Parser produces two outputs - trees with traces (with null elements) and without traces (null elements are removed). Examples of prepossessed trees, with trace and without trace, for English and Chinese are given in Figures \ref{ptb-examples} and \ref{ctb-examples}, respectively.

The removal of null elements in both Chinese and English is very similar; the Berkeley Neural Parser simply removes the null element, and leaves its parent node as a single branching constituent. This removal sometimes violates the syntactic integrity of the clause, but creates a tree without such a null element.
An example for English is given in Figure \ref{ptb-examples}, in which the VP, "to see if advertising works" requires a null subject  to represent the subject of the embedded clause to maintain its integrity. In removing this element, the Berkeley Neural Parser creates a sentence composed of just a VP, without a subject.
In the Chinese sentence given in Figure \ref{ctb-examples}, the embedded clause \begin{CJK}{UTF8}{gbsn}{从波黑撤军}\end{CJK} \textit{cóng bō hēi chèjūn} (`withdraw troops from Bosnia') has an arbitrary null subject, marked by the null element *PRO*. In order to remove this null element, the Berkeley Neural Parser creates an IP clause consisting of only a VP, without an NP subject. This is almost identical to the method used to remove the null subject in the earlier English sentence. 

Besides the deletion of null elements, a difference between sentences with and without trace in both English and Chinese is that sentences with trace contain extra information about each node. For instance, "NP-SBJ" is written rather than just "NP", which is what is written on trees without traces. This difference causes trees with traces to contain more information about each constituent; namely, what role they perform within the clause they are a part of. When trees are generated without trace, this information is not kept track of. Hence, the second labels are removed.

%While the Berkeley Neural Parser handles Chinese and English sentences, Our study team worked towards developing an implementation based off of this preexisting implementation to have the same effect on Korean sentences.

%An example for removal of Korean null elements is given in Figure 8. In that sentence, the null element from the pro-drop situation, labelled *pro*, is removed by having the S it is part of branch directly into the VP, so that it does not have an NP. The null operator labelled *op* is removed similarly, by removing the WHNP from the S that contains it, so that the S branches directly into the next S.

\section{Restoring Null Elements}
%explain what split is

Null element restoration has historically been approached in several ways -- namely through a rule based, pattern matching, or statistical approach. Much of this study has been focused on recovering null elements in the PTB, but there has been studies also for the CTB and KTB, especially in the restoration of *PRO* and *pro*, as these elements have been seen as particularly relevant for machine translation tasks. Our study has been focused on replicating rule based approaches for the PTB, and generating novel rules to expand a rule based approach to the CTB. Furthermore, we seek to expand the set of available approaches to null element restoration by running neural experiments in the PTB, CTB, and KTB. 

\subsection{Rules-based approach for PTB}
Campbell (2004) proposed an algorithm for restoring null elements by following a set of rules for each type of null element to deterministically insert them back into the trees. The rules used by Campbell proved to be highly effective, and had a strong theoretical basis in Government Binding Theory. Campbell argues that due to the fact that null elements result from the conscious implementation of linguistic rules by annotators, a rule based approach is particularly relevant for this task. The rules drawn from Campbell are detailed in Table~\ref{rules-based-approach-ptb}.
Our implementation has focused on revisiting and replicating Campbell's approach, since many of the details and code used are not accessible. By doing so, we hope to provide a more accessible baseline for future efforts to replicate linguistically oriented approaches to null element recovery.

\begin{table*}[!ht]
    \centering
\resizebox{\textwidth}{!}
{\footnotesize
\begin{tabularx}{\textwidth}{l X} \hline
\textsc{NP*} & If a node $X$ is a passive VP and has no complement, then insert NP* before all of its post-modifiers and any post-modifying dangling PP. However, if X is a non-finite clause and has no subject, then insert NP-SBJ* after all of its pre-modifiers. \\ \hdashline
\textsc{null complementizer} &  %{\color{red}it should be descriptive as before... }
If a node X is an SBAR and is not itself a complementizer, X has a child (Y) that is not a WHNP, and the siblings and parent nodes of X are not both NP, insert 0 to the left of Y.
%\begin{itemize}
%    \item Node $X$ is an SBAR and is not itself a complementizer
 %   \item X has a child node Y that is not a WHXP
  %  \item The parent node and sibling node of X are both not NP % or The parent node and sibling nodes of X are all not NP? 
%\end{itemize} 
\\\hdashline
\textsc{WHNP/WHADVP} & 
%\begin{itemize}
%    \item Node X is an SBAR and is not itself a complementizer
 %   \item X has a child node Y that is not a WHXP
  %  \item The parent node and sibling node of X are both NP % or The parent node and sibling nodes of X are all NP?
%\end{itemize}
If Node X is an SBAR and is not itself a complementizer, X has a child node Y that is not a WHXP, and both the parent node and sibling node of X are NP, check if the head of the parent of X belongs in one of the following categories: reason(s), way(s), time(s), day(s), place(s). If it does, then insert WHADVP to the left of Y. Otherwise, insert WHNP to the left of Y.\\\hdashline
\textsc{*U*} & Insert *U* as a sibling to the right of any QP that has a \$ child node followed by at least one CD child node. \\\hdashline
\textsc{WH-trace} & If a node X is an SBAR and it has a complementizer child node Y as well as a WHXP child node W, then find the trace(W) in Y. \\\hdashline
\textsc{Trace (W) in X} & 
To insert trace, there are eight cases to be considered, as follows. (1) If X has conjuncts then trace(W) should be found in the last conjunct instead. (2) If X has a PP child node with no object and W is a WHNP, then *T* should be inserted to the right of P. (3) If X = S, X is not a subject, and W is a WHNP, then *T* should be inserted as the last pre-modifier of X. (4) If X contains a VP, then trace(W) should be found in the VP. (5) If X contains an ADJP or clausal complement Y and W is a WHNP, then trace(W) should be found in the ADJP instead. (6) If W is a WHNP and has an infinitival relative clause R as a sibling, X is a VP and has an object NP, and the subject of R is an empty node E, then *T* should be inserted as the last pre-modifier of R, and E should be deleted. (7) If W is a WHNP, then *T* should be inserted as the first post-modifier of X. Finally, (8) if none of the above cases is true, then *T* should be inserted as the last post-modifier of X. \\\hline 

    \end{tabularx}
}
    \caption{A description of rules used to insert null elements into PTB sentences.}
    \label{rules-based-approach-ptb}
\end{table*}

\subsection{Rules-based approach for CTB}
In expanding a rules based approach to the CTB, we based our rules on information drawn from the CTB bracketing guidelines, as well as observations made in the CTB data set. One challenge faced was distinguishing between *pro* and *PRO* as these null elements have particularly similar distributions. The majority of both *pro* and *PRO* are accounted by the same pattern, (IP (NP (-NONE- *PRO/pro*)) (VP)). To differentiate between these two, we based our rules on patterns which included a wider amount of the tree found by Chung and Gildea (2010). This rule, however, does not have a strong theoretical basis, since it effectively works by identifying common patterns and inserting null elements, rather than by identifying a more general linguistic rule. Further work could be done to examine a more linguistically oriented approach to differentiating *pro* and *PRO* in null element recovery. For example, future approaches could use semantic role labelling to identify verb types that are common in control constructions, and use this information to differentiate *PRO* and *pro.* The rules implemented in our algorithm are detailed below in Table~\ref{rules-based-approach-ctb}. 

\begin{table*}[!ht]
    \centering
\resizebox{\textwidth}{!}
{\footnotesize    
\begin{tabularx}{\textwidth}{l X} \hline 
\textsc{*OP* and *T* in relative clause} & 
Consider a node X which is a CP, has a parent W that is a CP, and Child Y that is an IP. If Y has no daughter that is an NP, insert *T* in the subject position of the IP. Also insert a WHNP with a null *OP* as a parent to X. If Y does have an NP as a daughter, check the right daughter of IP, Z, which is a VP. If Z has a VP as a daughter, insert a *T* as an adjunct to Z. Also insert a WHPP with a null *OP* as a parent to X. Finally, if the above two rules do not apply and Z does not have an NP as a daughter, insert a *T* in the object position as the right daughter to Z and insert a WHNP with a null *OP* as a parent to X. 
\\ \hdashline

\textsc{*PRO* (control constructions)} &  
If any of the following patterns occur: ( VP VV NP ( IP VP ) ), 
( VP VV ( IP VP ) ), 
( CP ( IP VP ) DEC ), 
( PP P ( IP VP ) ), 
( LCP ( IP VP ) LC ), 
Insert (NP (-NONE- *PRO*)) as the left daughter of IP.

\\\hdashline

\textsc{*pro*} & 
If any of the following patterns occur: ( CP ( IP VP ) DEC ), 
( VP VV ( IP VP ) ), 
( LCP ( IP VP ) LC ), 
( IP IP PU ( IP VP ) PU ), 
( TOP ( IP VP PU ) ), 
Insert  (NP (-NONE- *pro*)) as the left daughter of the rightmost IP in the given pattern.

\\\hdashline

\textsc{*RNR*} & 
Consider a node X, which is a QP, with daughters Y and Z which are also both QPs. If Y has a left daughter, CD or OD, and no daughter labelled CLP, and Z has two daughters labelled CD or OD and CLP, insert (CLP (-NONE- *RNR*)) as a daughter to Y.
OR
Consider a node X, which is a VP, with daughters Y and Z which are also both VPs. If Y has no daughter labelled NP, and Z has a daughter labelled NP, insert (NP (-NONE- *RNR*)) as a daughter to Y.
\\\hline 

    \end{tabularx}
}
    \caption{A description of rules used to insert null elements into CTB sentences.}
    \label{rules-based-approach-ctb}
\end{table*}

\subsection{Language agnostic neural approach}

Syntactic information about a sentence may be represented in various non-linear formats, such as hierarchical tree structures or the CoNLL-U format. Linearization is the process of converting a non-linear structure into a linear sequence. The linear sequence can be represented in a single line, but maintains the syntactic information encoded in the non-linear structure. The purpose of lineariaztion is to allow for sequence-to-sequence neural network models to perform constituency and dependency parsing as a translation task. 
A simple method of linearizing a constituency tree is to perform a depth-first traversal of the tree and to order the nodes accordingly. Parentheses and POS tags can be added to preserve the hierarchical information in the linear format \citep{vinyals-etal-2015-grammar}.

\begin{figure*}
    \centering
\resizebox{\textwidth}{!}{    
\footnotesize    
\begin{tabularx}{\textwidth}{r|X} \hline 
\textsc{source}     &  (\textsc{top} (\textsc{s} (\textsc{np} ... )\textsc{np} (\textsc{vp} VBD (\textsc{np} (\textsc{np} NN )\textsc{np} (\textsc{sbar} (\textsc{whnp} WDT )\textsc{whnp} (\textsc{s} (\textsc{vp} MD (\textsc{vp})\textsc{vp} )\textsc{vp} )\textsc{s} )\textsc{sbar} )\textsc{np} )\textsc{vp} SFN )\textsc{s} )\textsc{top}
\\ \hdashline
\textsc{target}     &  (\textsc{top} (\textsc{s} (\textsc{np} ... )\textsc{np} (\textsc{vp} VBD (\textsc{np} (\textsc{np} NN )\textsc{np} (\textsc{sbar} (\textsc{whnp} WDT )\textsc{whnp} (\textsc{s} \textbf{(\textsc{np} *T* )\textsc{np}} (\textsc{vp} ... )\textsc{vp} )\textsc{s} )\textsc{sbar} )\textsc{np} )\textsc{vp} SFN )\textsc{s} )\textsc{top}\\\hline 
\end{tabularx}
}
    \caption{Example of the linearization dataset}
    \label{linearization-data}
\end{figure*}

We adapt linearization for null element restoration by translating the parse tree without null elements into a parse tree with null elements, as shown in Figure~\ref{linearization-data} where we convert it into a tree with a POS label as a terminal node. Since punctuation including initial and final punctuation marks in the Penn English  treebank should bet attached one level down to the highest level of \textit{labelled brackets} \citep{bies-EtAl:1995}, we add POS labels for punctuation, introduced in the Penn Korean Treebank, such as SFN for sentence-ending markers and SLQ for left quotation markers \citep{han-han:2001}. The Penn Chinese Treebank uses PU as a POS label for punctuation marks.

We utilize \textsc{t5}, or Text-To-Text Transfer Transformer \citep{rothe-etal-2021-simple}, a unified framework for NLP tasks that converts all tasks into a text-to-text format, for neural experiments.
We used the \texttt{T5-small} model, which is a smaller variant with approximately 60 million parameters with hyperparameters in Table~\ref{hyperparameters}.

\begin{table}[!ht]
    \centering
{\footnotesize    
    \begin{tabular}{r|l} \hline 
% hyperparameters = {
    'learning\_rate':&  1e-4,\\
    'batch\_size':& 16,\\
    'num\_epochs':& 10,\\
    'max\_length':& 512,\\
    'max\_new\_tokens':& 3500, \\
    'model\_name':& 'google-t5/t5-small’ \\ \hline 

    \end{tabular}
}
    \caption{Hyperparameters, used with RTX 4090 24GB VRAM and 80GB RAM}
    \label{hyperparameters}
\end{table}

\section{Experiments and Results}

\subsection{Evaluation methodology}
To evaluate the results of our experiments, we re-implemented Johnson's approach \citep{johnson-2002-simple}, which involves applying Parseval exclusively to nodes that constitute null elements. 
In particular, we implemented a novel approach based on \texttt{jp-evalb} \citep{park-etal-2024-jp,jo-park-park-2024-evalb} by first aligning terminal nodes, as opposed to the traditional \texttt{evalb} which requires consistent tokenization and sentence boundaries. 
Our evaluation metric does not consider non-terminal nodes; this is because we have taken a post-processing approach which involves inserting null elements into sentences which otherwise are identical to the gold trees. It is therefore assumed that if the null element is inserted in the correct location, the remaining bracketing is correct. Furthermore, our evaluation metric does not consider functional labels, as our goal was to restore null elements rather than to restore all functional labels.

For evaluation of the neural approach, since it uses a linearization format that excludes any non-null terminal nodes, additional processing is needed to recover the original format from the linearization dataset before it can be evaluated. This recovery has been done by inserting dummy words, meaning the original words are not restored. It should be noted that in the evaluation of the neural approach, some sentences were omitted from the calculation of the result due to limitations of the seq2seq model's ability to retain the original sentence's contents or to generate well-bracketed sentences. Evaluation can then be performed in the same method described previously in this section.

\subsection{Results}
The results for our experiments, as well as relevant past work, is summarized in Table~\ref{null-elements-insertion-results}, reported in terms of F1 scores.
We report our results using a rule-based approach for English and Chinese, and a neural approach for English, Chinese, and Korean. Given the small number of sentences (about 5000), we did not include rule-based results for Korean, as it was difficult to generalize rules for restoring null elements in Korean. 

\begin{table*}[!ht]
    \centering
\resizebox{\textwidth}{!}{    
\footnotesize{
    \begin{tabular}{ r c | ccccccc cc |c   } \hline 
     & & *T* & * & *U* & *0* & *PRO* & *pro* & *op* & *RNR* & *?* & Average \\ \hline
\citep{campbell-2004-using} & English  & 91.90  & 97.50 & 98.60 & 94.80  & - & - & - & - & - & 93.70 \\
\citep{johnson-2002-simple}  & English  & 85.90 & 88.00 & 95.00 & 94.00  & - & - & - &  - & - &  N/A \\
\citep{chung-gildea-2010-effects} & Chinese  & - & - & - & -  & 62.00 & 31.00 & - & - &  &  N/A \\ \hline 
Rules-based & English & 56.68 & 28.47 & 93.90 & 87.06  & - & - & - & - & - & 64.02 \\ 
Rules-based & Chinese & 84.03 & - & - & - & 71.49 & 39.08 & 91.92 & 78.57 & - & 80.00 \\ \hdashline

seq2seq & English (Without labels) & 88.49 & 83.24 & 99.09 & 96.73 & 86.50 & - & - &  & 64.43 & 88.67 \\ 
seq2seq & Chinese (Without labels) & 86.10 & - & - & - & 82.03 & 63.79 & 95.24 & 66.67 & - & 85.38 \\ 
seq2seq & Korean (Without labels) & 82.76 & - & - & -  & - & 91.61 & 92.14 &  & - &  88.79 \\ \hdashline

seq2seq & English (With labels) & 91.34 & 86.47 & 99.61 & 97.31 & 88.26 & - & - &  & 65.79 & 90.94 \\ 
seq2seq & Chinese (With labels) & 85.48 & - & - & -  & 79.86 & 64.22 & 92.27 & 72.73 & - & 83.76 \\ 
seq2seq & Korean (With labels) & 82.28 & - & - & - & - & 92.11 & 92.67 &  & - &  88.98 \\ \hline 
%Korean  & 736 & N/A & N/A & N/A & N/A & 950 & 656 & N/A & 11 & 2353 \\
   %& 31.28\% & N/A & N/A & N/A & N/A & 40.37\% & 25.88\% & N/A & 0.43\% & 100\% \\\hline 
    \end{tabular}
}
}
\caption{Results for null elements insertion, F1 scores reported.}  \label{null-elements-insertion-results}
\end{table*}

\subsection{Discussion}
Previous works have had difficulty with reliably restoring null elements in Chinese, particularly in the case of recovering *PRO* and *pro*. Our rule based implementation for Chinese Chinese yielded an average F1 score of 80.00, which is an improvement over Chung and Gildea's pattern-matching approach. The average F1 score would likely be greatly increased by improved scores for *PRO* and *pro* which are more conducive to approaches that combine lexical information with syntactic. This is because there are certain verb types that generate control constructions in Chinese, and identifying these specific verbs may allow a system to differentiate *PRO* from *pro*. For example, future approaches to Chinese may consider a hybrid which uses statistical features to capture relevant lexical information along with a syntactic rule based approach where it is effective. 

Our neural experiments began from linearization data, which means that they would have no access to lexical information. Despite this they achieved better results for *PRO* and *pro* compared to the rule-based approach and previous efforts \citep{chung-gildea-2010-effects}, which suggests that there are syntactic patterns that predict these elements, but may be difficult to identify via conventional means. 
% For the other null element types in Chinese, the rules based approach {\color{red}outperformed the neural experiment for all null elements other than *pro,* *PRO* and *T*, but had comparable results for *T*, (84.03, and 85.64 respectively) KX:UPDATE!}.  This is further evidence that the best approach for Chinese may be a hybrid approach that recovers most null elements in a rule-based way, but uses some sort of statistical approach for recovering *PRO* and *pro*. 
Furthermore, the neural experiments outperformed our rule-based approach approach for every null element other than *RNR*, which has a very clear-cut structure under which it is inserted.
In English, neural results still lag behind past approaches, such as Campbell's work, which may give credence to the claims made by Campbell that a rules-based approach is the most effective strategy to recover null elements in English. The neural experiments lagged behind Campbell's work for *T* and *, while outperforming it for *U* and 0. However, considering the relatively simple seq2seq model runs in this paper, further work in applying neural experiments to recover null elements may achieve results that are more comparable to past approaches in English.
% Interestingly, the neural experiments had comparable results in Chinese (80.51) and English (83.86), but much greater results for Korean (91.35). This may be because Korean has fewer null element types making them easier to predict. The null element types also have more distinct syntactic contexts; for example, this may contribute to the significant difference in recovering *pro* in Chinese (64.22) and *pro* in Korean (94.69). 
The neural experiments for Korean showed significantly better results for pro (92.11 and 92.14) compared to Chinese (64.22 and 63.79). We believe this is because the model does not need to differentiate between PRO and pro as it does in Chinese.
Some precision and recall was lost on neural experiments, as certain non-existent null elements were hallucinated through the neural experiment. Preventing these hallucinations may also work to increase the viability of a neural approach to null element recovery. In English, the additional tokens *ICH* and *EXP* are added, which are not explored in any previous works.
% which are not found in the Penn Treebank guide, nor are they explored in any previous works. 
The results for these elements were thus not included, but their insertion has decreased the accuracy of our neural experiments overall as well. Future works may aim to exclude these elements if their inclusion is deemed not useful.
Lastly, including functional labels has improved constituency parsing results for Korean \citep{chung-post-gildea:2010:SPMRL}, but not for French \citep{park:2018:TALN:Parsing}. This explains our neural results with and without functional labels, where only English and Korean showed improvement, while Chinese did not.
Linearization experiments are not based on CFG rules where adding functional labels affect the size and shape of CFG rules. We leave it as future work to investigate how restoring null elements in the parse tree (including constituency parsing) would be affected by the different sequence when functional labels are added.

\section{Conclusion}

% Ultimately, a rules-based approach may be the best suited to the task of null element recovery, considering that past rule-based work outperformed neural experiments, and our rule-based implementation for Chinese outperformed the neural implementation for most null element types in Chinese. 
This study has improved on previous works by exploring more null elements in a rule-based approach for Chinese, as well as providing a basis for a language-agnostic neural approach for future works to expand on, which also explores null elements ignored by previous works.
Ultimately, a neural approach may be the best suited to the task of null element recovery, considering that the neural experiments had comparable results to previous work for English and to our rule-based implementation for Chinese in elements other than *pro* and *PRO*.
Although future work could improve upon our rule-based Chinese results to surpass our neural experiments, rule-based approaches are difficult to implement, and require a theoretical background in a given language. In contrast, language agnostic neural experiments can be more easily implemented across several languages and robustly handle null elements that previous work ignored. Furthermore, considering our use of a relatively simple architecture, future work in using seq2seq to recover null elements may produce results that compare or outperform rules-based approaches. 
% This is especially true for Korean, since the achieved results, even in our relatively simple experiments, were comparable to past work in null element recovery for English. 
More broadly, a neural approach to null element recovery is particularly useful for any languages that lack an existing language specific rule-based approach and for languages with a small number of null element types like Korean. With further work to improve on the baseline results for neural experiments set in this paper, neural results could be viable cross-linguistically as a general approach to null element recovery.

\section*{Limitations}
This study has primarily focused on the use of syntactic context to predict null elements. Future work interested in a linguistically oriented approach to null element recovery may benefit from the use of Semantic Role Labelling, or other tools that capture lexical information, to generate rules to predict some null element types. Furthermore, our neural experiments used linearized data which intentionally excluded lexical information, so that we could identify the efficacy of of neural networks in predicting null elements based on syntactic context. Future approaches may similarly enhance results by including lexical information.

\end{document}